\pdfoutput=1
\documentclass[11pt,a4paper]{article}
\usepackage[margin=1in]{geometry}

\usepackage[]{acl}

\usepackage{times}
\usepackage{latexsym}

\usepackage[T1]{fontenc}
\usepackage[utf8]{inputenc}

\usepackage{microtype}

\usepackage{graphicx}
\usepackage{listings}
\usepackage{enumitem}

\providecommand{\wgs}[1]
{\noindent
  \begin{small}	
  \textbf{\textit{Relevant UniDive working groups:~}} #1
  \end{small}
}

\title{{Annotating Constructions with UD: \\ 
the experience of the Italian Constructicon}}

 \author{Ludovica Pannitto \\
  University of Bologna \\
  \texttt{\small{ludovica.pannitto@unibo.it}} \\ 
  \textbf{Beatrice Bernasconi} \\
  University of Torino \\ 
  \texttt{\small{beatrice.bernasconi@unito.it}}   \\ \And
  Lucia Busso \\
  Aston University \\
  \texttt{\small{l.busso@aston.ac.uk}} \\ 
 \textbf{Flavio Pisciotta} \\
  University of Salerno \\
\texttt{\small{fpisciotta@unisa.it}} \\ \And
Giulia Rambelli \\
University of Bologna\\
\texttt{\small{giulia.rambelli4@unibo.it}}\\
\textbf{Francesca Masini} \\
University of Bologna \\
\texttt{\small{francesca.masini@unibo.it}}
 }

\begin{document}
\maketitle

\wgs{WG1, WG2}

\section{Constructicography}

In constructionist theory, the constructicon is the inventory and network of constructions (henceforth, \textit{cxns}) makes up the full set of linguistic units in a language. Constructicography \cite{lyngfelt2018constructicography} then needs to deal with two interconnected aspects, namely:
\begin{itemize}[noitemsep,nolistsep]
    \item the \textit{dictionary of cxns}: a linguistically-motivated repository of entries including what we consider necessary and sufficient information to represent cxns for multiple purposes (language documentation and learning, etc.);
    \item the identification and annotation of the actual occurrences of cxns (i.e., constructs) in texts. 
\end{itemize}

We argue that it is necessary to tackle them at the same time and cross-linguistically to properly set the path for organic resource development. 

Cxns are defined to be conventional form-function pairings at varying levels of complexity and abstraction \cite{goldberg1995constructions,goldberg2006constructions}, ranging from morphology to syntax to discourse. 
As a matter of fact, linguistic resources are mostly available for the morphosyntactic level of analysis and the UD infrastructure represents the \textit{de facto} standard in this area, providing a framework which has been employed to represent the most diverse set of languages (both typologically as well as in terms of language variation - i.e., genre, spoken vs. written, code switching, etc.).
At the same time, various initiatives exist to augment the UD proposed annotation to include morphological analysis \cite{guillaume2024joint,kyjanek2020universal}, multiword patterns (as in PARSEME, \citealt{parseme}), semantic role labeling (as in the Universal PropBank project \citealt{jindal-EtAl:2022:LREC}). A proposal also exists for adding an annotation layer for construction-like information \cite{weissweiler-etal-2024-ucxn}. 

%We therefore argue that this makes it the ideal test ground to start developing a shared representation not only to annotate constructions on existing resources, but to develop constructicons themselves. To this end, we propose a new sub-task for WG2 that tackles these issues, building on the experience we honed developing the Italian Constructicon. 

%\section{AdoC: the Italian Constructicon}
\section{The Italian Constructicon}

%We present a pilot formalization developed, within the \textit{Adopt a Construction} (AdoC) project \cite{masini-etal-2024-adoc}, for the Italian constructicon. 
The formalization we introduce consists of two distinct but interconnected data structures, namely:
\begin{itemize}[noitemsep,nolistsep]
\item {\textbf{the graph of cxns} (\texttt{GCxn})}, a repository of constructions expressed by a set of constraints and organized into a directed graph;
\item {\textbf{a body of annotated examples} incrementally built by annotating the occurrences of a specific cxn (i.e., constructs) in texts.}
\end{itemize}

% and can be translated as a series of grew queries. Constructions, which are nodes in the graph, are connected by links that represent connections holding within the graph (these can be either vertical/inheritance links or horizontal/similarity links);

% # GREW

\subsection{GCxn and the \texttt{conll-c} format}

\begin{figure*}
\footnotesize{
\begin{lstlisting}
# cxn-id = 68
# cxn-name = saltare fuori che V
# function = ref:D is found out unexpectedly
# vertical_links = 
# horizontal_links = 167

ID  UD.FORM   LEMMA    UPOS              FEATS                  HEAD    DEPREL
A   _         saltare  VERB              Number=Sing|Person=3    0      root
B   fuori     fuori    ADV               _                       A      advmod
C   che       che      SCONJ             _                       D      mark
D   _          _       VERB, ADJ, NOUN   _                       A      csubj

REQUIRED   WITHOUT                SEM.FEATS   SEM.ROLES   ADJACENCY   IDENTITY
1          CHILDREN:DEPREL=nsubj   _           _             _           _
1          _                       _           _             _           _
1          _                       _           _             _           _
1          _                       _          Eventuality    _           _
\end{lstlisting}
}

\captionof{lstlisting}{CoNLL-C annotation for the cxn \textit{salta fuori che} V `it turns out that V' (lit. `it jumps out that V'). The columnar format was split in two lines for space reasons.}
\label{fig:conll-c}
\end{figure*}

Within the graph, each cxn is identified by a unique identifier and a human-intelligible (i.e., conventional or easy to grasp) name, and is described through a set of fields stored in \texttt{yaml} format.

Most fields are machine-readable, coherently with the aim to define, as much as possible, the properties of a cxn through a UD-compatible formalization. Formally speaking, a cxn is defined as a set of rooted, directed, acyclic, labeled graphs (i.e., dependency trees), where nodes represent constraints on the cxn components (i.e., tokens and sub-token elements) and edges represent constraints on relations among tokens.
These constraint can be expressed in what we call \texttt{conll-c} format, a newly introduced convention to maximize compatibility with \texttt{conll-u}. 
Cxns formalized in \texttt{conll-c} can be translated into a set of \texttt{grew} queries and be used to automatically match UD-parsed sentences in order to find cxn instances.

The \texttt{conll-c} format mimics \texttt{conll-u} with respect to file format (plain UTF-8 text files, containing either tokens, blank lines or cxn-level comments starting with hash). Cxns, just like sentences, consist of one or more word lines.
% \footnote{In \texttt{conll-u}, lines refer to \textit{words}, but our notion of token can apply below the word level (i.e., to prefixes or morphological roots in the case of morphological constructions).}
For clarity reasons, we refer to each element of a cxn as a “token”, but our notion of token can apply below the word level (i.e., to affixes or roots in the case of morphological constructions). Listing \ref{fig:conll-c} shows an example of \texttt{conll-c} annotation of the mirative construction \textit{salta fuori che} V `it turns out that V' (id $=$ 68).

%Annotations of cxns are encoded in plain text files (UTF-8) that contain three types of lines: blank lines, token lines containing the annotation of a component/token/node in 13 fields separated by single tab characters, and 
Cxn-level comments (starting with hash, \#), include:
\begin{itemize}[noitemsep,nolistsep]
    \item the name and identifier of the cxn (\texttt{\# cnx\_id} and \texttt{\# cxn\_name}); 
    \item a short natural-language string representing its function (\texttt{\# cxn\_function}); it can include references to specific tokens (in the form of ref:tok\_ID) that allow to link the function definition to actual cxn instances; 
    \item a space separated list of cxn ids to which it is vertically linked (parent cxns, \texttt{\# vertical\_links});
    \item a space separated list of cxn ids to which it is horizontally linked (sibiling cxns\texttt{\# horizontal\_links});
    \item other optional metadata that describe holistic features of the cxn.
\end{itemize}

\begin{figure}[h]
\tiny{
\begin{lstlisting}
# sent_id = 2_Paisa_FP06072024
# source = http://hdl.handle.net/20.500.12124/3 paisa.raw.utf8/7404393 .
# text = [omitted for space reasons]

1   Poi       poi     ADV    4    advmod    _
2   ci        ci      PRON   4    expl      CXN=345:B
3   siamo     essere  AUX    4    aux       _
4   messi     mettere VERB   0    root      CXN=345:A
5   a         a       ADP    6    mark      CXN=345:C
6   parlare   parlare VERB   4    xcomp     CXN=345:D
7   ...       ...     PUNCT  9    punct     _
8   e         e       CCONJ  9    cc        _
9   salta     saltare VERB   4    conj      CXN=68:A
10  fuori     fuori   ADV    9    advmod    CXN=68:B
11  che       che     SCONJ  14   mark      CXN=68:C
12  era       essere  AUX    14   cop       _
13  davvero   davvero ADV    14   advmod    _
14  Chris     Chris   PROPN  9    ccomp     CXN=68:D
15  Squire    Squire  PROPN  14   flat:name _
16  !         !       PUNCT  4    punct     _
17  !         !       PUNCT  4    punct     SpaceAfter=No
\end{lstlisting} }
\captionof{lstlisting}{Annotation for the sentence \textit{Poi ci siamo messi a parlare...e salta fuori che era davvero Chris Squire!!} `Then we got to talking...and it turns out it was really Chris Squire!!!'. For space reasons, only some fields are shown.}
\label{fig:conll-u}
\end{figure}

Fields express the constraints that constructs need to fulfill in order to be considered instances of the construction.
More specifically:
Field 1 (\texttt{ID}) is compulsory and expresses the token index by means of uppercase alphabetic letter starting at A for each new cxn. For tokens containing sub-word  (i.e., morphological) information, \texttt{ID} is formed by the letter of the derived token and an integer starting at 1 for each subcomponent of the token (e.g., A*1, A*2, etc.).
Fields 2-7 can be left unspecified (\texttt{\_}) if no constraint applies, or can be filled with comma separated values compatible with UD notation. More specifically, \texttt{FORM} and \texttt{LEMMA} can also be expressed by means of regular expressions. In case of subcomponents of the token:
\begin{itemize}[noitemsep,nolistsep]
    \item allomorphs are lemmatized as their morpheme (e.g., in Italian all allomorphs of \textit{in-} such as \textit{ir-}, \textit{il-}, \textit{im-}... are lemmatized as \textit{in-});
    \item \texttt{UPOS} is set as \texttt{BMORPH} (\textit{bound morpheme});
    \item as a convention, the full word is the \texttt{HEAD} of the principal subword element;
    \item we extend the tagset with the following labels for morphological relations: \texttt{root/m} (the stem for derivation, the head for compounding, or the first element in case of coordinate compounds), \texttt{der/m} (links derivational affixes to the stem), \texttt{case/m} (links complement to head in subordinate compounds), \texttt{mod/m} (links attribute to head in attributive compounds), \texttt{conj/m} (links constituents in coordinate compounds).
\end{itemize}

Field 8 (\texttt{REQUIRED}) encodes whether the token has to necessarily appear in every construct (value \texttt{1}) or is optional (value \texttt{0}).
Field 9 (\texttt{WITHOUT}) stores values that have to be excluded from matching. It is expressed as a pipe separated list of \texttt{FIELD\_NAME=value}. A special case is for expressing that no other token can depend on the token in question through a specific relation, in which case we use the syntax \texttt{CHILDREN=deprel}.
Fields 10-11 encode constraints on the ontological class of the token (in case of Ns and ADJs, from Open Multilingual WordNet\footnote{\url{https://omwn.org/doc/topics.html}}), its Aktionsart (in case of verbs), and its semantic role (adapted from VerbNet\footnote{\url{https://verbs.colorado.edu/verbnet/}} and the Unified Verb Index\footnote{\url{https://uvi.colorado.edu/}}).\footnote{Obviously, not all of these are matchable in a UD-environment, but they still serve the purpose of building a machine-readable version of a cxn. For the same reason, cxns are linked to \url{https://comparative-concepts.github.io/cc-database/}.}. 
Field 12 (\texttt{ADJACENCY}) refers to the linear adjacency of tokens within the sentence. It is filled with either \_ for \textit{any} or the \texttt{ID} of the left-adjacent to the one in object.
Lastly, field 13 (\texttt{IDENTITY}) encodes constraints on co-indexification (e.g., if token with \texttt{ID} C has to have the same \texttt{FORM} of token with \texttt{ID} A, the value will be \texttt{FORM=A})

\subsection{Annotated Examples}

Examples are maintained in \texttt{conll-u} format. In the \texttt{MISC} column, the value \texttt{CXN=cxn\_id:ID} is added for relevant tokens. As shown in Listing \ref{fig:conll-u}, a single example can be annotated for more than one construction: in this case, both \textit{salta fuori che} V `it turns out that V' (id $=$ 68) and \textit{mettersi a} V `begin to V' (id $=$ 345) are annotated.
Moreover, metadata is enriched with a \texttt{\# source} field which links the example to the original source, in order to automatically update UD-derived annotation and properly split into train, development, and test files.

\section{Building the Constructicon}
The Constructicon can be populated both manually and semi-automatically. Each cxn is formalized or revised by a linguist that compiles the \texttt{yaml} description, there including also the \texttt{conll-c} formalization.
The cxn is then pushed to GCxn where, through CI/CD tools, it will be translated into a series of \texttt{grew} queries which are used to automatically retrieve sentences that might contain instances of the cxn. These (or a subset) are manually revised by a linguist in order to remove false positives.
An algorithm will be then run to check graph consistency and update vertical links. Once the position in the graph is set, all sentences associated with the previously existing cxns can be ideally re-analized and enriched in light of information coming from the newly added cxn (i.e., if cxn N is parent of cxn M in the graph and an example has been added for cxn M, than that example can be automatically annotated as an instance of cxn N, too).

\bibliography{anthology,custom}
\bibliographystyle{acl_natbib}

\end{document}